\documentclass[journal,transmag]{IEEEtran}
\usepackage{amsmath,amssymb,amsfonts}
\usepackage{amsthm}
\usepackage{algorithm}
\usepackage{algorithmic}
\usepackage{graphicx}
\usepackage{textcomp}
\usepackage{xcolor}
\usepackage{comment}
\usepackage{url}
\usepackage{cite}
\usepackage{booktabs}
\usepackage{hyperref}

\begin{document}

\title{Bodhi VLM: Privacy-Alignment Modeling for Hierarchical Visual Representations in Vision Backbones and VLM Encoders via Bottom-Up and Top-Down Feature Search}

\author{\IEEEauthorblockN{Bo Ma\IEEEauthorrefmark{1}, \IEEEauthorrefmark{2}
Wei Qi Yan\IEEEauthorrefmark{1}
Jinsong Wu\IEEEauthorrefmark{3},\IEEEauthorrefmark{4}
}
\IEEEauthorblockA{
\IEEEauthorrefmark{1} Auckland University of Technology, Auckland 1024 NZ,\\
\IEEEauthorrefmark{2} Resideo Technologies, Inc., Austin, TX 78759, USA,\\
\IEEEauthorrefmark{3} Guilin University of Electronic Technology, Gui Lin, China, \\
\IEEEauthorrefmark{4} Universidad de Chile, Santiago, Chile, 
}

}

\markboth{}%
{Bodhi VLM: Privacy-Alignment Modeling for Hierarchical Visual Representations}

\IEEEtitleabstractindextext{%
\begin{abstract}
Learning systems that preserve privacy often inject noise into hierarchical visual representations; a central challenge is to \emph{model} how such perturbations align with a declared privacy budget in a way that is interpretable and applicable across vision backbones and vision--language models (VLMs). We propose \emph{Bodhi VLM}, a \emph{privacy-alignment modeling} framework for \emph{hierarchical neural representations}: it (1) links sensitive concepts to layer-wise grouping via NCP and MDAV-based clustering; (2) locates sensitive feature regions using bottom-up (BUA) and top-down (TDA) strategies over multi-scale representations (e.g., feature pyramids or vision-encoder layers); and (3) uses an Expectation-Maximization Privacy Assessment (EMPA) module to produce an interpretable \emph{budget-alignment signal} by comparing the fitted sensitive-feature distribution to an evaluator-specified reference (e.g., Laplace or Gaussian with scale $c/\epsilon$). The output is reference-relative and is \emph{not} a formal differential-privacy estimator. We formalize BUA/TDA over hierarchical feature structures and validate the framework on object detectors (YOLO, PPDPTS, DETR) and on the \emph{visual encoders} of VLMs (CLIP, LLaVA, BLIP). BUA and TDA yield comparable deviation trends; EMPA provides a stable alignment signal under the reported setups. We compare with generic discrepancy baselines (Chi-square, K-L, MMD) and with task-relevant baselines (MomentReg, NoiseMLE, Wass-1). Results are reported as mean$\pm$std over multiple seeds with confidence intervals in the supplementary materials. This work contributes a learnable, interpretable modeling perspective for privacy-aligned hierarchical representations rather than a post hoc audit only.
Source code: \href{https://github.com/mabo1215/bodhi-vlm.git}{Bodhi-VLM GitHub repository}
\end{abstract}
}

\maketitle
\IEEEdisplaynontitleabstractindextext
\IEEEpeerreviewmaketitle

\section{Introduction}
\label{sec:intro}

Privacy-preserving learning systems often inject noise into hierarchical visual representations (e.g., feature pyramids in detectors or vision-encoder layers in VLMs). A central question for \emph{neural network modeling} is how to characterize and interpret the alignment between observed layer-wise perturbations and a declared privacy budget $\epsilon$ under an evaluator-specified reference mechanism (e.g., Laplace or Gaussian with scale $c/\epsilon$). We propose \emph{Bodhi VLM}, a \emph{privacy-alignment modeling} framework for hierarchical visual representations that: (1) \emph{models} sensitive feature structure in a layer-wise manner via bottom-up (BUA) and top-down (TDA) grouping over multi-scale features; and (2) produces an interpretable \emph{budget-alignment signal} via an Expectation-Maximization Privacy Assessment (EMPA) module that compares the fitted sensitive-feature distribution to the reference. The output is reference-relative and does not constitute formal differential privacy certification or estimation of effective $\epsilon$.

BUA and TDA traverse hierarchical feature structures (e.g., feature pyramid networks in detectors, or the visual encoder layers in CLIP~\cite{radford2021learning} and LLaVA~\cite{liu2023llava}), producing paired sensitive and non-sensitive feature groups per layer using NCP and MDAV-based clustering; EMPA then uses these groups in an EM-like procedure to output a discrepancy signal. The framework does not modify the underlying architecture but provides a learnable, interpretable view of how privacy-related noise is distributed across layers and how it aligns with a declared budget.

Contributions: (1) We formalize BUA and TDA for sensitive feature search over FPN and VLM vision backbones using NCP and MDAV-based clustering. (2) We introduce EMPA as a feature-level budget-alignment (discrepancy) estimator and integrate it into Bodhi VLM. (3) We show how Bodhi VLM applies to the \emph{vision towers} of representative VLMs (Section~\ref{subsec:vlm}); extending the assessment to full cross-modal fusion and language-decoding pathways remains future work. (4) We report experiments on MOT20 and COCO and design an experimental protocol for comparison (BUA vs.\ TDA, EMPA vs.\ Chi-square, K-L, MMD, and task-relevant baselines MomentReg, NoiseMLE, Wass-1). (5) We compare Bodhi VLM with representative VLM privacy methods (ViP, DP-Cap, DP-MTV, VisShield) and with distribution-level and task-specific baselines (Section~\ref{subsec:related-comparison}); we demonstrate the framework on CLIP, LLaVA, and BLIP vision encoder features (Section~\ref{sec:vlmcompare}).

\paragraph{Scope and relevance to TNNLS.}
The core contribution of this work is \emph{modeling} for hierarchical neural representations: we propose a formalization of layer-wise sensitive-feature structure (BUA/TDA), a grouping mechanism (NCP/MDAV) over multi-scale visual features, and an interpretable budget-alignment estimator (EMPA). This is a contribution to \emph{learning systems and representation structure}, not only to applied privacy evaluation. The problem is situated within neural network architectures (vision backbones and VLM encoders) and is of interest to the TNNLS readership because it addresses how to model and interpret privacy-related perturbations in hierarchical representations produced by learned vision and vision--language models.

\section{Related Work}
\label{sec:related}

\subsection{Differential Privacy and Privacy Auditing}
\label{subsec:related-dp}
Differential privacy (DP)~\cite{dwork2008differential} provides a rigorous notion of privacy loss via a budget $\epsilon$. Assessing whether a trained model satisfies a given $\epsilon$ has been studied from both theoretical and empirical angles. One-line or few-run privacy auditing~\cite{narayanan2021rethinking,jagielski2022differentially} aims to estimate the effective $\epsilon$ from model behavior; our EMPA layer provides a complementary, feature-level assessment that consumes the outputs of BUA/TDA. Hitaj et al.~\cite{hitaj2017deep} use deep generative models to study information leakage in collaborative learning. Agrawal and Aggarwal~\cite{agrawal2001design} quantify privacy in data mining. DP for deep learning~\cite{abadi2016deep} and for federated settings~\cite{mcmahan2017communication} are standard baselines; Bodhi VLM does not replace training-time DP but evaluates whether the observed noise aligns with the declared budget.

\subsection{Hierarchical and Multimodal Representations}
\label{subsec:related-repr}
Feature pyramid networks (FPN)~\cite{lin2017feature} offer multi-scale representations used in object detection; we leverage their hierarchical structure for BUA and TDA. Vision-language models (VLMs) combine a visual encoder with a language model: CLIP~\cite{radford2021learning} aligns image and text embeddings via contrastive learning; LLaVA~\cite{liu2023llava} projects visual features into the LLM space for instruction following; BLIP~\cite{li2022blip} unifies encoding and decoding for vision-language understanding and generation. Recent VLMs such as Qwen2-VL~\cite{wang2024qwen2vl} and InternVL~\cite{chen2024internvl} scale to 4K resolution and support video understanding, expanding the scope of multimodal tasks. These architectures expose layer-wise visual features (ViT blocks or CNN stages), to which BUA and TDA can be applied without modifying the model.

\subsection{Privacy in Vision and Vision-Language Models}
\label{subsec:related-vlm}
Privacy for VLMs has gained attention with several distinct approaches. \textbf{ViP}~\cite{yu2024vip} trains differentially private vision transformers via masked autoencoders and DP-SGD on LAION400M ($\epsilon{=}8$), achieving 55.7\% ImageNet linear probing. \textbf{DP-Cap}~\cite{sander2024dpcap} improves upon ViP using image captioning for multimodal DP representation learning, reaching 65.8\% ImageNet-1K under $\epsilon{=}8$ on LAION-2B. \textbf{DP-MTV}~\cite{ngong2026dpmtv} enables many-shot multimodal in-context learning with $(\varepsilon,\delta)$-DP by aggregating demonstrations into compact task vectors, preserving 50\% accuracy on VizWiz at $\varepsilon{=}1.0$. \textbf{VisShield}~\cite{chen2025visshield} adopts a de-identification paradigm: it uses instruction-tuned VLMs to locate and mask Protected Health Information (PHI) in medical images via OCR. \textbf{Membership inference} on VLMs~\cite{hu2025vlmmia} reveals that instruction-tuning data is vulnerable to attacks achieving AUC $>0.8$ with few samples. Bodhi VLM complements these works: whereas ViP/DP-Cap/DP-MTV focus on training with DP and VisShield on de-identification, Bodhi VLM assesses whether the \emph{observed} privacy noise aligns with a declared budget, providing a feature-level audit tool that can be applied to any VLM vision tower.

\subsection{External Comparison Methods Used in This Work}
\label{subsec:related-comparison}
Our core innovation is \emph{hierarchical sensitive-feature modeling}, \emph{grouping mechanisms} (BUA/TDA with NCP/MDAV), and \emph{budget-alignment estimation} (EMPA) for neural representations---i.e., a modeling and representation-level contribution rather than a purely applied privacy-evaluation tool. We position Bodhi VLM against the following external methods and baselines in our experiments (Section~\ref{sec:exp}). \textbf{VLM privacy methods:} ViP~\cite{yu2024vip}, DP-Cap~\cite{sander2024dpcap}, DP-MTV~\cite{ngong2026dpmtv}, and VisShield~\cite{chen2025visshield} are contrasted with Bodhi VLM in Table~\ref{tab:vlmprivacy}; they represent training-time DP or de-identification rather than post hoc budget assessment, so the comparison is qualitative (paradigm and compatibility with our framework). \textbf{Distribution-level and task-relevant baselines:} We compare EMPA with generic discrepancy measures (Chi-square, K-L divergence, MMD) and with \emph{task-specific budget-alignment} baselines: \emph{MomentReg} (regression of declared $\epsilon$ from feature moments and layer-wise dispersion), \emph{NoiseMLE} (maximum-likelihood under the evaluator-specified additive noise family), and \emph{Wass-1} (first-order Wasserstein distance between perturbed and reference sensitive-feature distributions). These appear in Tables~\ref{tab:rmse}, \ref{tab:vlmemparesults}, and in the supplementary materials. \textbf{Random-partition baseline:} On synthetic hierarchical features we add a \emph{random partition + EMPA} baseline (same sensitive-set size as BUA/TDA but partition chosen uniformly at random) to show that structured grouping (BUA/TDA) yields a more stable and interpretable budget-alignment signal than a blind partition (Figure~\ref{fig:script_empa}). Together, these external comparison methods allow readers to assess Bodhi VLM relative to both prior VLM privacy work and to generic distributional and non-structured baselines.

\subsection{Expectation-Maximization and Microaggregation}
\label{subsec:related-em}
The EM algorithm~\cite{dempster-laird-rubin-77} and its extensions~\cite{em-converge,greff2017neural} inspire our EMPA formulation. Microaggregation (e.g., MDAV~\cite{soria2014enhancing}) and NCP-based grouping are used in both BUA and TDA to form sensitive/non-sensitive clusters. Similar ideas appear in $k$-anonymity and $l$-diversity~\cite{machanavajjhala2007diversity} for tabular data; we adapt them to continuous feature vectors in neural backbones.

\section{Method}
\label{sec:method}

\subsection{Formal Problem Definition}
\label{subsec:problem}

We distinguish three conceptual components: (1) the \emph{privacy mechanism} that perturbs data or features; (2) \emph{sensitive feature localization}, which identifies layer-wise features corresponding to a sensitive label set $S$ (or $S_\epsilon$ when we emphasize the budget under which it is used); and (3) \emph{reference-perturbation assessment}, which measures how well the observed perturbed features match an \emph{evaluator-defined reference} distribution. Bodhi VLM addresses (2) and (3) only; it does not certify formal end-to-end differential privacy guarantees.

\paragraph{Inputs and output (reference-relative).}
Let $D$ be an image dataset; $f$ the original (unperturbed) layer-wise features; $\tilde{f}$ the observed (perturbed) features; $S$ a set of sensitive labels; $\epsilon$ a declared budget; and $\mathcal{M}_{\mathrm{ref}}(\epsilon; \eta)$ an \emph{evaluator-defined reference mechanism} with family $\eta$ (e.g., Laplace with scale $c/\epsilon$). The assessment is \emph{reference-relative}: given $(\tilde{f}, f, S, \epsilon, \mathcal{M}_{\mathrm{ref}})$, we compute a \emph{reference-perturbation discrepancy score} $\mathrm{BAS}_\epsilon = A(\tilde{f}, f, S, \epsilon; \mathcal{M}_{\mathrm{ref}})$ that compares the distribution of observed sensitive-group features to the distribution induced by $\mathcal{M}_{\mathrm{ref}}$ on the same subset. The output is a scalar discrepancy (and optionally weight-only bias or rMSE to that reference). We do \emph{not} claim that $\mathrm{BAS}_\epsilon$ estimates the effective $\epsilon$ of the true mechanism or provides formal DP certification; if the deployed mechanism differs from $\mathcal{M}_{\mathrm{ref}}$, the score reflects model mismatch as well as budget alignment.

\subsection{Preliminary and Notation}
\label{subsec:preliminary}

Let $D$ be an image dataset and $S_\epsilon$ the set of sensitive labels. A privacy-preserving pipeline processes $D$ and adds noise to sensitive regions. We assume a backbone that produces feature maps at layers $i=1,\ldots,n$, represented as layer-wise feature sets $\{X_i\}_{i=1}^{n}$. At each layer $i$, $\mathcal{G}^{(n)}_i$ denotes the non-sensitive group and $\mathcal{G}^{(s)}_i$ the sensitive group. Table~\ref{tab:notation} summarizes the main notation. The goal is to (1) partition features at each layer into $\mathcal{G}^{(n)}_i$ and $\mathcal{G}^{(s)}_i$ via BUA or TDA, and (2) from these groups, compute a budget-alignment signal via EMPA (a bias and/or rMSE to a reference). Bodhi VLM achieves this via BUA, TDA, and EMPA.

\begin{table*}[t]
\caption{Main notation.}
\label{tab:notation}
\centering
\begin{tabular}{ll}
\hline
Symbol & Meaning \\
\hline
$D$ & Image dataset \\
$S$, $S_\epsilon$ & Sensitive label set (subscript $\epsilon$ when emphasizing budget) \\
$\mathcal{M}_{\mathrm{ref}}(\epsilon;\eta)$ & Evaluator-defined reference mechanism (family $\eta$) \\
$f$, $\tilde{f}$ & Original and perturbed layer-wise features \\
$X_i$ & Set of feature vectors at layer $i$ (e.g., tokens/patches/locations) \\
$\mathcal{G}^{(n)}_i$, $\mathcal{G}^{(s)}_i$ & Non-sensitive and sensitive groups at layer $i$ \\
$t$, $u$, $v$ (EMPA) & Raw sensitive data, interference (noise), observed data ($v=t+u$) \\
$s(x;S_\epsilon)$, $\tau_i$ & Sensitivity score and layer-wise threshold \\
$\lambda_k$, $\mu_k$, $\sigma_k^2$ (EMPA) & Mixture weights, means, diagonal variances in EMPA \\
\hline
\end{tabular}
\end{table*}

\subsection{Sensitive Concepts and Scores}
\label{subsec:sensitive-protocol}

In all experiments, $S_\epsilon$ denotes a \emph{task-defined} set of sensitive concepts rather than a quantity learned from $\epsilon$ itself. For MOT20, we set $S_\epsilon=\{\texttt{person}\}$ for all tested budgets. For COCO, we consider a person-only'' setting with $S_\epsilon=\{\texttt{person}\}$ and an expanded traffic-sensitive'' setting with $S_\epsilon=\{\texttt{person},\texttt{car},\texttt{truck},\texttt{bus}\}$; the subscript in $S_\epsilon$ therefore indicates the sensitive set used under the declared budget, not that the semantic set is generated by $\epsilon$.

For a layer-wise feature vector $x_{i,j}\in X_i$, the sensitivity score $s(x_{i,j};S_\epsilon)$ depends on the backbone family. For detector backbones, we first map $x_{i,j}$ to the corresponding region in the detector head and define $s(x_{i,j};S_\epsilon)$ as the maximum detector-head confidence among classes in $S_\epsilon$ for that region. For VLM vision encoders (CLIP, LLaVA, BLIP), we compute
\[
s(x_{i,j};S_\epsilon)
=
\max_{c\in S_\epsilon}
\mathrm{sim}\bigl(h_i(x_{i,j}),q_c\bigr),
\]
where $h_i(\cdot)$ maps the local visual feature to the vision encoder's embedding space, $q_c$ is the text-embedding prototype for concept $c$, and $\mathrm{sim}$ is cosine similarity. Unless otherwise stated, we set the per-layer threshold $\tau_i$ as the 90th percentile of $s(\cdot;S_\epsilon)$ within layer $i$, so that approximately the top 10\% most sensitive features (under this score) are assigned to $\mathcal{G}^{(s)}_i$.

\subsection{Model Structure}
\label{subsec:structure}

\begin{figure*}[t]
\centering
\includegraphics[width=0.8\textwidth]{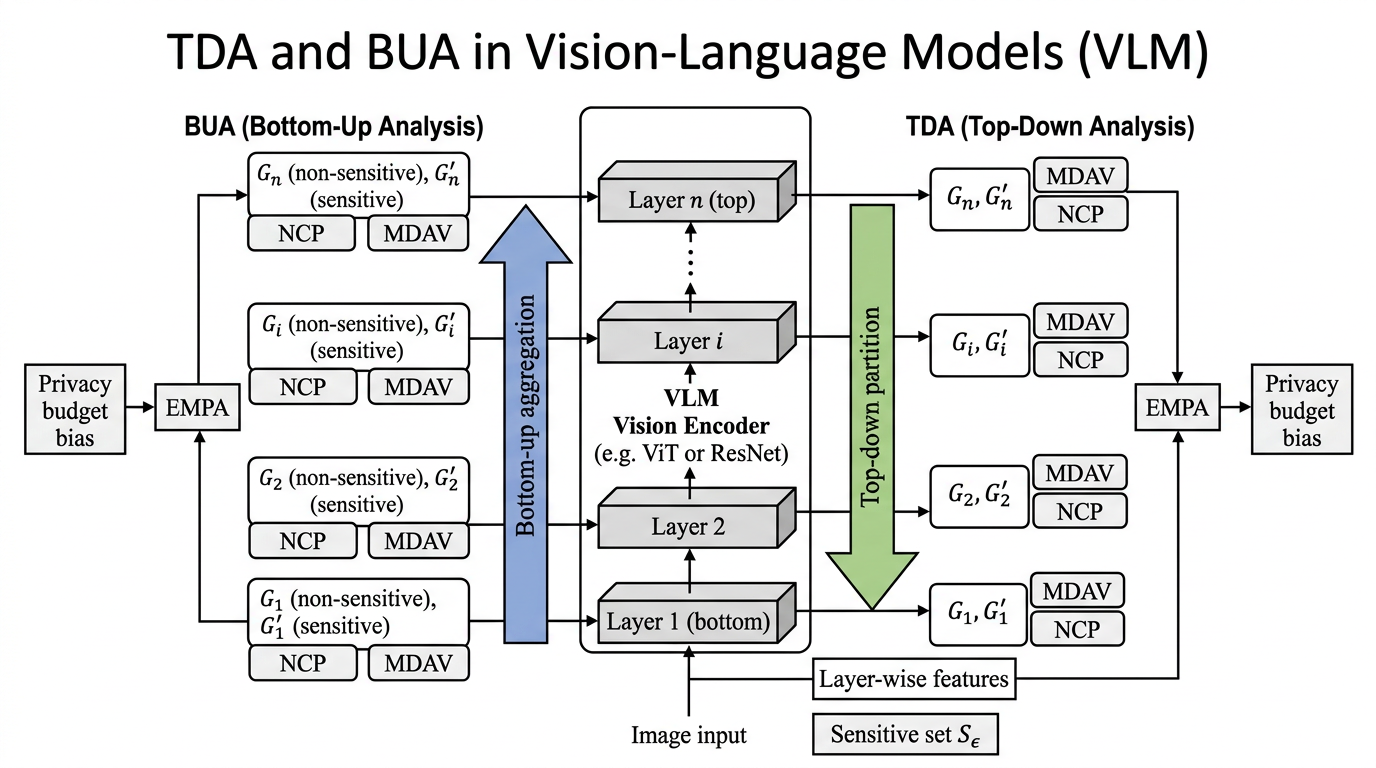}
\caption{TDA and BUA in VLM: top-down and bottom-up feature search over the vision encoder layers (e.g., ViT blocks); layer-wise groups $\mathcal{G}^{(n)}_i$, $\mathcal{G}^{(s)}_i$ feed EMPA for feature-level budget-alignment assessment.}
\label{fig:tda}
\end{figure*}

The Bodhi VLM framework comprises three main components, as illustrated in Fig.~\ref{fig:bua} and Fig.~\ref{fig:tda}: (1) \textbf{Bottom-Up Strategy (BUA)} aggregates low-level features upward through the backbone layers, using NCP and MDAV to form sensitive and non-sensitive groups per layer. (2) \textbf{Top-Down Strategy (TDA)} partitions high-level features downward and links layer-wise groups via set intersections and NCP. (3) \textbf{Expectation-Maximization Privacy Assessment (EMPA)} takes the layer-wise partitions $\{\mathcal{G}^{(s)}_i\},\{\mathcal{G}^{(n)}_i\}$ from either BUA or TDA, together with $\epsilon$ and $S_\epsilon$, and estimates whether the observed perturbation aligns with the declared budget via an EM-inspired procedure, outputting a discrepancy signal (bias). In this work, Bodhi VLM is validated primarily on the visual encoders (or vision towers) of representative vision-language models under an explicit perturbation mechanism; extending the same budget-alignment assessment to full cross-modal fusion and language-generation pathways remains future work. We detail each component next.

\subsection{Bottom-Up Strategy (BUA)}
\label{subsec:bua}

BUA starts from the bottom layer of the backbone and iterates upward over layer-wise feature sets $\{X_i\}_{i=1}^{n}$, where each $X_i$ contains feature vectors (e.g., per-patch tokens for ViT, or per-location vectors for CNN feature maps). At each layer $i$, BUA outputs a partition of $X_i$ into a sensitive group $\mathcal{G}^{(s)}_i$ and a non-sensitive group $\mathcal{G}^{(n)}_i$ based on a sensitivity score $s(x;S_\epsilon)$ and a threshold $\tau_i$:
\begin{equation}
s(x_{i,j};S_\epsilon) \in [0,1],
\label{eq:sensitivity-score}
\end{equation}
\begin{equation}
\mathcal{G}^{(s)}_i
:=
\left\{
x_{i,j} \in X_i \;:\; s(x_{i,j};S_\epsilon)\ge \tau_i
\right\},
\qquad
\mathcal{G}^{(n)}_i = X_i \setminus \mathcal{G}^{(s)}_i.
\label{eq:sensitive-partition}
\end{equation}
We use MDAV~\cite{soria2014enhancing} as a microaggregation operator for grouping (e.g., to enforce a minimum group size and stabilize the partition), and NCP as a quantitative penalty to measure within-group generalization (defined below). Algorithm~\ref{alg:bua} gives a layer-wise specification.

\paragraph{NCP and MDAV definitions.}
For a cluster (group) $C$ of $d$-dimensional feature vectors, we define the normalized certainty penalty (NCP) as
\begin{equation}
\operatorname{NCP}(C)
:=
\sum_{a=1}^{d}
w_a \, \operatorname{penalty}_a(C),
\label{eq:ncp}
\end{equation}
with
\begin{equation}
\operatorname{penalty}_a(C)
:=
\frac{
\max_{x \in C} x^{(a)} - \min_{x \in C} x^{(a)}
}{
R_a
},
\label{eq:ncp-penalty}
\end{equation}
where $w_a \ge 0$ are attribute weights and $R_a$ is the range (or a robust range estimate) of dimension $a$. We apply MDAV to the feature vectors in $X_i$ (or an explicit transform $\phi(X_i)$ if used in implementation) to form microaggregated groups before computing $s(\cdot;S_\epsilon)$ and the partition in Eq.~\eqref{eq:sensitive-partition}; we state the chosen space (raw feature vs.\ transformed) explicitly in experiments.

\begin{figure*}[t]
\centering
\includegraphics[width=0.9\textwidth]{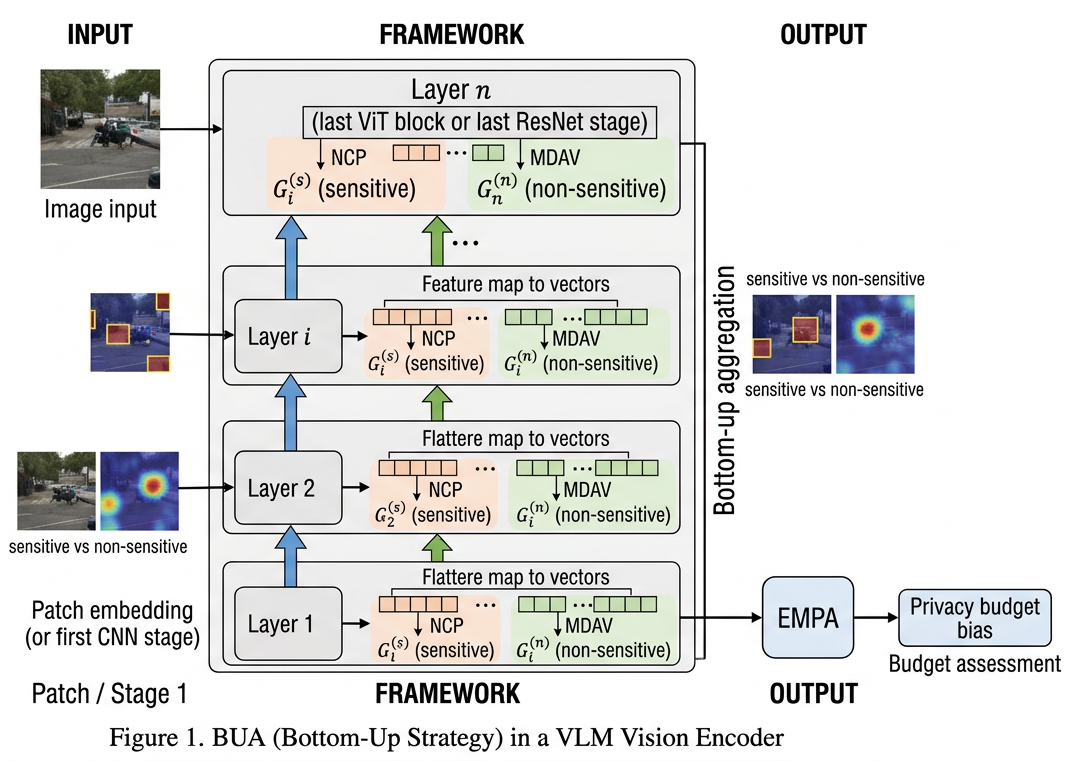}
\caption{BUA in a VLM vision encoder: image input and patch embedding (or first CNN stage) at the bottom; bottom-up aggregation through Layer~1 to Layer~$n$ with NCP/MDAV partitioning into sensitive $\mathcal{G}^{(s)}_i$ and non-sensitive $\mathcal{G}^{(n)}_i$ at each layer; output to EMPA for privacy budget assessment.}
\label{fig:bua}
\end{figure*}

\paragraph{BUA in VLMs.}
In a VLM, the ``bottom'' layer $i=1$ is the patch embedding or the first CNN stage (e.g., CLIP ViT patch projection, LLaVA's ViT~\cite{liu2023llava}, or BLIP's ResNet stage~1). BUA then moves upward through the vision encoder: for a ViT, layer $i$ is the output of the $i$-th transformer block (before or after the MLP); for a ResNet-based encoder, $i$ indexes the residual stages. At each $i$, the feature map is flattened into vectors (e.g., per-patch or per-spatial-location), and NCP and MDAV operate on these vectors with the same $S_\epsilon$ used for the rest of the pipeline. The upward aggregation thus builds a hierarchy of sensitive/non-sensitive groups that aligns with the VLM's internal representation levels, from local patches (low-level) to global semantics (high-level), and the layer-wise partitions $\{\mathcal{G}^{(s)}_i\},\{\mathcal{G}^{(n)}_i\}$ are passed to EMPA for budget assessment. Fig.~\ref{fig:bua} illustrates BUA in a VLM vision encoder: from image input and patch embedding (or first CNN stage), features are flattened to vectors at each layer; NCP and MDAV partition them into sensitive $\mathcal{G}^{(s)}_i$ and non-sensitive $\mathcal{G}^{(n)}_i$ groups along the bottom-up aggregation, and the layer-wise partitions are passed to EMPA for privacy budget assessment.

\begin{algorithm}[t]
\caption{Bottom-Up Strategy (BUA)}
\label{alg:bua}
\begin{algorithmic}[1]
\REQUIRE Layer-wise feature sets $\{X_i\}_{i=1}^{n}$; sensitive concept set $S_\epsilon$; thresholds $\{\tau_i\}_{i=1}^{n}$
\ENSURE Sensitive groups $\{\mathcal{G}^{(s)}_i\}_{i=1}^{n}$ and non-sensitive groups $\{\mathcal{G}^{(n)}_i\}_{i=1}^{n}$
\FOR{$i=1,2,\dots,n$}
  \STATE (Optional) Apply MDAV microaggregation to $X_i$ (in the chosen space) to obtain stable groups.
  \STATE Compute sensitivity scores $s(x;S_\epsilon)$ for all $x \in X_i$.
  \STATE Partition $X_i$ into $\mathcal{G}^{(s)}_i=\{x \in X_i: s(x;S_\epsilon)\ge \tau_i\}$ and $\mathcal{G}^{(n)}_i = X_i \setminus \mathcal{G}^{(s)}_i$.
  \STATE (Optional) Evaluate within-group dispersion using $\operatorname{NCP}(\mathcal{G}^{(s)}_i)$ and $\operatorname{NCP}(\mathcal{G}^{(n)}_i)$ for auditing/stability.
\ENDFOR
\STATE \textbf{return} $\{\mathcal{G}^{(s)}_i\}_{i=1}^{n}$ and $\{\mathcal{G}^{(n)}_i\}_{i=1}^{n}$
\end{algorithmic}
\end{algorithm}

Figure~\ref{fig:bua} illustrates BUA in a VLM vision encoder: the image input is processed by patch embedding (or the first CNN stage), yielding an initial sensitive vs.\ non-sensitive visualization; the framework then aggregates features bottom-up through Layer~1 to Layer~$n$, with each layer producing flattened feature vectors partitioned into $\mathcal{G}^{(s)}_i$ (sensitive) and $\mathcal{G}^{(n)}_i$ (non-sensitive) via NCP and MDAV; the resulting layer-wise groups feed into EMPA for budget assessment and privacy budget bias estimation.

\subsection{Top-Down Strategy (TDA)}
\label{subsec:tda}

TDA starts from the top layer and proceeds downward. Like BUA, it produces a sensitive/non-sensitive partition at each layer, but additionally uses inter-layer links to propagate the grouping from higher to lower layers. Because $\mathcal{G}^{(s)}_i$ and $\mathcal{G}^{(s)}_{i-1}$ generally lie in different feature spaces (different dimensions or coordinate systems), we first map higher-layer elements to the lower layer via a correspondence operator $\Pi_{i \to i-1}: X_i \to X_{i-1}$ (e.g., nearest spatial/patch correspondence or identity when dimensions match). We then define the top-down links:
\begin{equation}
L^{(s)}_i = \Pi_{i \to i-1}(\mathcal{G}^{(s)}_i) \cap \mathcal{G}^{(s)}_{i-1},
\qquad
L^{(n)}_i = \Pi_{i \to i-1}(\mathcal{G}^{(n)}_i) \cap \mathcal{G}^{(n)}_{i-1},
\label{eq:td-links}
\end{equation}
and use these links to refine the lower-layer grouping (e.g., by adjusting thresholds $\tau_{i-1}$ or by reweighting sensitivity scores for elements connected to high-level sensitive groups). Algorithm~\ref{alg:tda} specifies the layer-wise operations. We use MDAV for microaggregation and NCP for within-group dispersion as in BUA; importantly, we avoid reusing EMPA variables ($u,v$) in the algorithmic notation.

\paragraph{TDA in VLMs.}
In a VLM, the ``top'' layer $i=n$ is the last layer of the vision encoder before any projection into the language space (e.g., the last ViT block in CLIP~\cite{radford2021learning} or LLaVA, or the final global-pooled features in BLIP~\cite{li2022blip}). TDA then walks backward: $i=n-1,n-2,\ldots,1$. At each step, the same MDAV/NCP logic partitions the current layer's feature vectors into $\mathcal{G}^{(n)}_i$ and $\mathcal{G}^{(s)}_i$, and the intersections $\mathcal{G}^{(n)}_i \cap \mathcal{G}^{(n)}_{i-1}$ and $\mathcal{G}^{(s)}_i \cap \mathcal{G}^{(s)}_{i-1}$ tie high-level semantic groups to the lower layers that support them. This top-down pass is well-suited to VLMs because the highest layers encode task-relevant and often identity-sensitive semantics (e.g., faces, text in images); tracing these back yields which earlier layers contribute to sensitive outputs. The resulting per-layer partitions are again fed into EMPA together with $\epsilon$ and $S_\epsilon$.

\begin{algorithm}[t]
\caption{Top-Down Strategy (TDA)}
\label{alg:tda}
\begin{algorithmic}[1]
\REQUIRE Layer-wise feature sets $\{X_i\}_{i=1}^{n}$; sensitive concept set $S_\epsilon$; thresholds $\{\tau_i\}_{i=1}^{n}$
\ENSURE Sensitive groups $\{\mathcal{G}^{(s)}_i\}_{i=1}^{n}$ and non-sensitive groups $\{\mathcal{G}^{(n)}_i\}_{i=1}^{n}$
\STATE Initialize top-layer grouping at $i=n$ by computing $s(x;S_\epsilon)$ for $x\in X_n$ and applying Eq.~\eqref{eq:sensitive-partition}.
\FOR{$i=n,n-1,\dots,2$}
  \STATE (Optional) Apply MDAV microaggregation to $X_{i-1}$ (in the chosen space).
  \STATE Compute scores $s(x;S_\epsilon)$ for all $x \in X_{i-1}$.
  \STATE Form a preliminary partition $(\mathcal{G}^{(s)}_{i-1},\mathcal{G}^{(n)}_{i-1})$ using Eq.~\eqref{eq:sensitive-partition}.
  \STATE Compute top-down links $L^{(s)}_i=\mathcal{G}^{(s)}_i \cap \mathcal{G}^{(s)}_{i-1}$ and $L^{(n)}_i=\mathcal{G}^{(n)}_i \cap \mathcal{G}^{(n)}_{i-1}$.
  \STATE Refine the lower-layer partition using the propagated links (e.g., prioritize elements in $L^{(s)}_i$ as sensitive).
\ENDFOR
\STATE \textbf{return} $\{\mathcal{G}^{(s)}_i\}_{i=1}^{n}$ and $\{\mathcal{G}^{(n)}_i\}_{i=1}^{n}$
\end{algorithmic}
\end{algorithm}

\subsection{Expectation-Maximization Privacy Assessment (EMPA)}
\label{subsec:empa}

EMPA is a \emph{reference-relative discrepancy} module: it takes the layer-wise groups from BUA or TDA, a declared budget $\epsilon$, and an \emph{evaluator-defined reference mechanism} $\mathcal{M}_{\mathrm{ref}}(\epsilon;\eta)$, and outputs an empirical discrepancy between the observed sensitive-feature distribution and the reference. EMPA is not a formal estimator of end-to-end differential privacy parameters.

\paragraph{Concrete instantiation and reference.}
We use a $K$-component diagonal-Gaussian mixture on sensitive-group feature vectors. Let $\mathcal{V}^{(s)}=\{v_j\}_{j=1}^{N_s}$ denote all feature vectors in sensitive groups across selected layers (after perturbation, optionally pooled to a common space). We model
\begin{equation}
p(v \mid \Theta_\epsilon)
=
\sum_{k=1}^{K} \lambda_k \,
\mathcal{N}\bigl(v;\mu_k,\mathrm{diag}(\sigma_k^2)\bigr),
\label{eq:empa-mixture}
\end{equation}
with $\Theta_\epsilon=\{\lambda_k,\mu_k,\sigma_k^2\}_{k=1}^K$, $\lambda_k\ge 0$, $\sum_{k=1}^K \lambda_k=1$. We fit $\Theta_\epsilon$ by standard EM: E-step responsibilities $\gamma_{jk} = \lambda_k \mathcal{N}(v_j;\mu_k,\mathrm{diag}(\sigma_k^2)) / \sum_{\ell=1}^K \lambda_\ell \mathcal{N}(v_j;\mu_\ell,\mathrm{diag}(\sigma_\ell^2))$, and M-step updates $N_k = \sum_j \gamma_{jk}$, $\lambda_k^{\mathrm{new}} = N_k/N_s$, $\mu_k^{\mathrm{new}} = (1/N_k)\sum_j \gamma_{jk} v_j$, $(\sigma_k^2)^{\mathrm{new}} = (1/N_k)\sum_j \gamma_{jk}(v_j-\mu_k)^2$.

The \emph{reference} $\Theta_\epsilon^{\mathrm{ref}}$ is obtained by the same fitting on sensitive-group features generated by $\mathcal{M}_{\mathrm{ref}}(\epsilon;\eta)$ (e.g., Laplace with $b=c/\epsilon$) on the same subset. Thus the score is explicitly \emph{reference-relative}; if the true mechanism differs from $\mathcal{M}_{\mathrm{ref}}$, the discrepancy reflects both budget alignment and mechanism mismatch. Robustness to $K \in \{2,4,8\}$, diagonal vs.\ full covariance, and to the choice of $\psi$ below is left for future work.

\paragraph{Discrepancy function.}
We define the \emph{reference-perturbation discrepancy score} (reported as budget-alignment score in tables) by a parameterization $\psi(\Theta)$ and $\ell_2$ distance:
\begin{equation}
\psi(\Theta) = \big[\lambda_1,\ldots,\lambda_K,\;\mu_1^\top,\ldots,\mu_K^\top,\;(\sigma_1^2)^\top,\ldots,(\sigma_K^2)^\top\big]^\top,
\end{equation}
\begin{equation}
\mathrm{BAS}_{\epsilon}
:=
\left\| \psi(\Theta_{\epsilon}) - \psi(\Theta_{\epsilon}^{\mathrm{ref}}) \right\|_2.
\label{eq:budget-alignment-score}
\end{equation}
In our implementation we use the full $\psi$ unless stated otherwise. When we focus on a weight-only view, we report
\begin{equation}
\mathrm{Bias}_{\epsilon}
:=
\left\|
\lambda_{\epsilon}
-
\lambda_{\epsilon}^{\mathrm{ref}}
\right\|_2,
\label{eq:bias}
\end{equation}
where $\lambda_\epsilon$ and $\lambda_\epsilon^{\mathrm{ref}}$ denote the weight vectors from $\Theta_\epsilon$ and $\Theta_\epsilon^{\mathrm{ref}}$. These quantities are empirical feature-level discrepancy scores under the chosen perturbation model and declared budget; they are not estimates of an effective differential-privacy parameter and should not be interpreted as such.

\paragraph{Primary and secondary outputs.}
The \emph{primary} output is the reference-perturbation discrepancy score $\mathrm{BAS}_\epsilon = \|\psi(\Theta_\epsilon)-\psi(\Theta_\epsilon^{\mathrm{ref}})\|_2$, which compares the fitted sensitive-feature distribution to the reference under the same declared $\epsilon$ and family $\mathcal{M}_{\mathrm{ref}}$. The quantity $\mathrm{Bias}_\epsilon = \|\lambda_\epsilon-\lambda_\epsilon^{\mathrm{ref}}\|_2$ is a weight-only diagnostic. Neither is a formal estimator of effective $\epsilon$; both are reference-relative empirical signals.

\subsection{Extension to Vision-Language Models}
\label{subsec:vlm}

Vision-language models (VLMs) such as CLIP~\cite{radford2021learning}, LLaVA~\cite{liu2023llava}, and BLIP~\cite{li2022blip} use a visual encoder (e.g., ViT or ResNet) that produces hierarchical or multi-layer features before fusion with text. We validate Bodhi VLM on the \emph{vision towers} of these VLMs; extending the assessment to full cross-modal fusion and language-decoding pathways remains future work. Without architectural change: (1) Treat the visual encoder as a backbone with layers $i=1,\ldots,n$ (e.g., ViT blocks or CNN stages). (2) Run BUA from the earliest layer upward or TDA from the last visual layer downward to obtain sensitive/non-sensitive groups $\mathcal{G}^{(n)}_i$, $\mathcal{G}^{(s)}_i$ per layer, using the same NCP and MDAV-based clustering over feature vectors and a given sensitive label set $S_\epsilon$. (3) Feed the resulting groups into EMPA to estimate the discrepancy signal between observed privacy noise and the declared budget $\epsilon$.

\paragraph{Layer mapping and feature extraction.}
For \textbf{CLIP} (ViT backbone), $i=1$ is the patch embedding output; $i=2,\ldots,n$ are the transformer block outputs (e.g., $n=12$ for ViT-B/32). Each layer's feature tensor has shape (patches, dim); we treat each patch (or pooled region) as one vector for NCP/MDAV. For \textbf{LLaVA}, the vision tower is typically a ViT (e.g., CLIP ViT-L/14); again $i=1$ is patch embedding and $i=2,\ldots,n$ are block outputs. The language model receives only the final layer (or a projected subset); Bodhi VLM runs BUA/TDA on all vision layers and does not modify the language branch. For \textbf{BLIP} with a ResNet image encoder, $i$ indexes the residual stages (e.g., $i=1,\ldots,4$); features are spatial maps that we flatten to vectors per location. In all cases, $S_\epsilon$ can be defined over visual concepts (e.g., person, vehicle) or over text labels if image-text pairs are available; the same grouping logic then identifies which layer-wise features align with those concepts.

\paragraph{Text and multimodal sensitivity.}
Text modalities can be incorporated by defining $S_\epsilon$ over joint image-text sensitive concepts (e.g., person IDs, license plates) and applying the same grouping on the visual branch; the language branch can be left unmodified or assessed separately. In this work, Bodhi VLM is validated primarily on the visual encoders (or vision towers) of representative vision-language models under an explicit perturbation mechanism. Extending the same budget-alignment assessment to full cross-modal fusion and language-generation pathways remains future work.

\section{Experiments}
\label{sec:exp}

\subsection{Experimental Setup}
\label{sec:expdesign}

\subsubsection{Datasets}
We use MOT20~\cite{dendorfer2020mot20} (150 frames, uniformly sampled from the designated train/val sequences as in the official split) and COCO~\cite{lin2014microsoft} (train2017/val2017 splits, $k$-fold with $k=10$ where applicable). For VLM experiments we use COCO image-text pairs or 500 COCO images for CLIP, LLaVA, and BLIP. Synthetic hierarchical features (4 layers, 200 samples, 8-D per layer, 30\% sensitive ratio) are used to validate the experimental protocol. Privacy budgets: $\epsilon \in \{0.1, 0.01, 0.001\}$.

\subsubsection{Evaluation Metrics}
We report \textbf{deviation} between original and privacy-noised features (BUA vs.\ TDA), \textbf{rMSE} for budget alignment (EMPA vs.\ Chi-square, K-L divergence, MMD), and \textbf{EMPA bias} (deviation of fitted mixture weights from uniform). For detector experiments we also report PSNR and SIFT-based metrics where applicable~\cite{cruz2012scale,sara2019image}. In the main tables, the \emph{target} of assessment is the \emph{observed feature perturbation} relative to a reference (e.g., Laplace) or to the declared budget; rMSE is the root mean squared error between the metric's output and this reference (or between noised and original features where stated). \textbf{Lower rMSE is better} when measuring budget-alignment error. We do not claim that these metrics estimate the effective $\epsilon$ of a formal DP mechanism; they provide empirical alignment signals for comparison.

\subsubsection{Baselines}
We compare EMPA with generic distributional baselines (Chi-square, K-L divergence, MMD) and with \emph{task-relevant budget-alignment} baselines: \emph{MomentReg} (regression of declared $\epsilon$ from feature moments and layer-wise dispersion), \emph{NoiseMLE} (MLE under the evaluator-specified additive noise family), and \emph{Wass-1} (first-order Wasserstein distance between perturbed and reference sensitive-feature distributions). Results for MomentReg, NoiseMLE, and Wass-1 are reported in the main tables where space permits and in the supplementary materials (Tables~S1--S2). For VLM privacy we contrast Bodhi VLM with ViP~\cite{yu2024vip}, DP-Cap~\cite{sander2024dpcap}, DP-MTV~\cite{ngong2026dpmtv}, and VisShield~\cite{chen2025visshield} (Table~\ref{tab:vlmprivacy}).

\subsubsection{Implementation Details}
Detectors: YOLOv4~\cite{bochkovskiy2020yolov4}, PPDPTS, MDCRF+YOLO, DETR~\cite{carion2020end}. VLMs: CLIP ViT-B/32~\cite{radford2021learning}, LLaVA-1.5 7B~\cite{liu2023llava}, BLIP (ResNet-50)~\cite{li2022blip}. We apply additive Gaussian or Laplace noise to sensitive regions/features. \textbf{Perturbation calibration used in controlled experiments:} for Laplace perturbation we use $u \sim \mathrm{Laplace}(0,b)$ with $b = c/\epsilon$, and for Gaussian perturbation $u \sim \mathcal{N}(0,\sigma^2 I)$ with $\sigma = c/\epsilon$, where $c$ is a fixed constant shared across all methods in a given experiment. This controlled setup is used only to generate a reference perturbation family for empirical auditing and is not claimed as a formal DP calibration. It is important to distinguish: the \textbf{declared budget} $\epsilon$ is the parameter under the (assumed) mechanism; the \textbf{noise scale} used in simulation ($c/\epsilon$) is a choice for controlled experiments; the \textbf{observed feature perturbation} is what we measure (e.g., deviation, rMSE). Our experiments evaluate budget-alignment under these controlled additive mechanisms and do not certify formal end-to-end DP for full VLM pipelines. BUA/TDA use NCP and MDAV with sensitive set $S_\epsilon$; EMPA runs E/M steps until convergence. 

\paragraph{Statistical reporting.}
Where multi-seed results are reported (e.g., synthetic experiments in Section~\ref{sec:scriptresults}, Table~\ref{tab:script}), we repeat the evaluation over five random seeds $\{0,1,2,3,4\}$ covering perturbation sampling, grouping, and EMPA initialization, and we report mean $\pm$ standard deviation. Detector and VLM tables (Tables~\ref{tab:rmse}, \ref{tab:vlmemparesults}) also use mean$\pm$std over the same five seeds; corresponding 95\% confidence intervals are computed from these multi-seed runs and provided in the supplementary materials.

\subsubsection{Reproducibility Configuration}
For clarity, Table~\ref{tab:repro_details} summarizes the main implementation choices and hyperparameters used in our experiments; these defaults are shared across detector and VLM vision-encoder settings unless otherwise specified.

\begin{table*}[t]
\centering
\caption{Reproducibility details for BUA/TDA/EMPA used in our experiments.}
\label{tab:repro_details}
\begin{tabular}{p{4cm}p{11cm}}
\hline
Item & Specification \\
\hline
Feature normalization & $\ell_2$ normalization per feature vector before MDAV/NCP (where applied) \\
MDAV group size & $k=8$ (minimum cluster size for microaggregation) \\
NCP weights $w_a$ & Uniform: $w_a=1/d$ for a $d$-dimensional feature \\
Range $R_a$ & Inter-quantile range $(q_{0.95}-q_{0.05})$ computed on the training split for each dimension \\
Sensitivity threshold $\tau_i$ & $90$th percentile of $s(\cdot;S_\epsilon)$ within layer $i$ (per Section~\ref{subsec:sensitive-protocol}) \\
BUA aggregation unit & Grouping computed per image; summary statistics averaged across images \\
TDA cross-layer mapping & Nearest spatial/patch correspondence between consecutive layers (same grid or nearest-neighbor in coordinates) \\
EMPA component number $K$ & $K=4$ diagonal-Gaussian components in all reported experiments \\
EMPA stopping criterion & Relative log-likelihood improvement $<10^{-5}$ or a maximum of 100 EM iterations \\
Reported values & Mean$\pm$std over 5 seeds for detector (Table~\ref{tab:rmse}), synthetic (Table~\ref{tab:script}), and VLM (Table~\ref{tab:vlmemparesults}); confidence intervals estimated from the multi-seed runs (see supplementary materials) \\
Hardware & NVIDIA GPUs (exact model and PyTorch version documented in the code release) \\
\hline
\end{tabular}
\end{table*}

\subsection{Overall Performance}
\label{sec:overall}

We summarize the main results below; detailed figures and tables follow in the next sub-subsections. Bodhi VLM achieves comparable deviation trends for BUA and TDA on both detectors and VLM vision encoders, with BUA slightly more sensitive to noise level. EMPA shows stable budget-alignment (rMSE and bias) across MDCRF, DETR, and PPDPTS, and on CLIP vision encoder features (Tables~\ref{tab:rmse}, \ref{tab:vlmemparesults}); BLIP results are omitted when the model is unavailable in the run.

\subsubsection{BUA vs.\ TDA}
\begin{figure}[t]
\centering
\includegraphics[width=0.9\columnwidth]{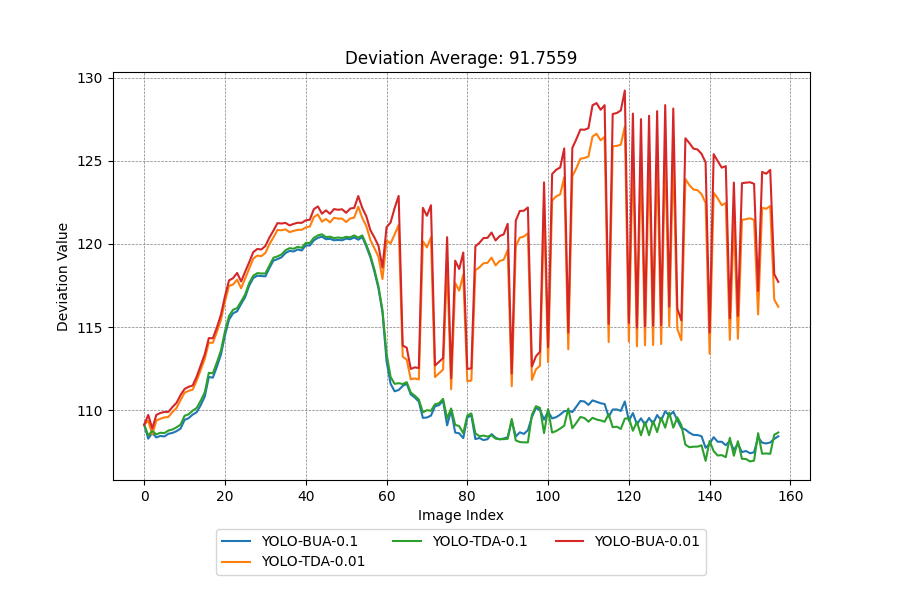}
\caption{BUA vs.\ TDA deviation on YOLO (MOT20).}
\label{fig:yolo}
\end{figure}

\begin{figure}[t]
\centering
\includegraphics[width=0.9\columnwidth]{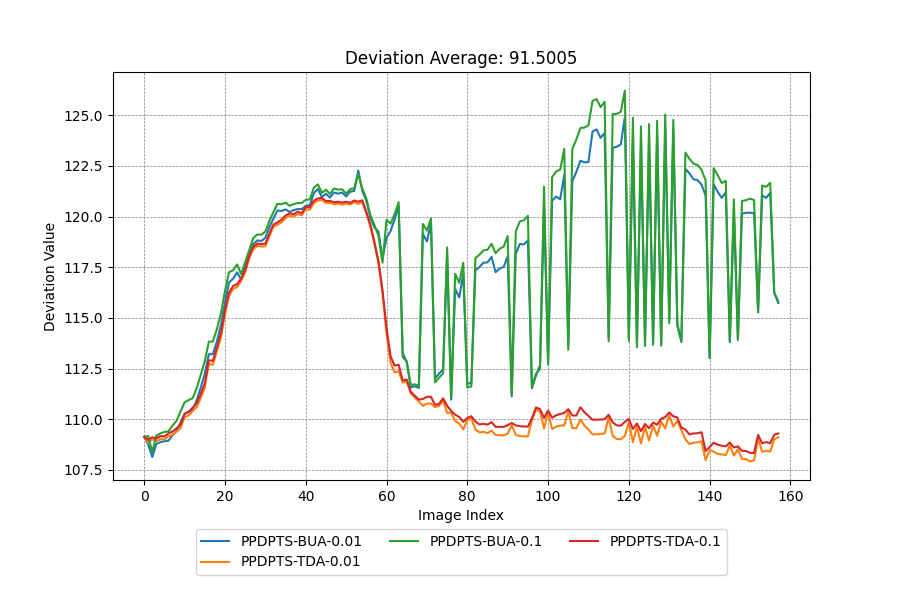}
\caption{BUA vs.\ TDA deviation on PPDPTS (MOT20).}
\label{fig:ppdpts}
\end{figure}

Figures~\ref{fig:yolo} and~\ref{fig:ppdpts} show the deviation between original and privacy-noised images (MOT20) under BUA and TDA with $\epsilon \in \{0.1, 0.01\}$. Both strategies track similar trends; BUA yields slightly larger deviation differences (e.g., max difference TDA vs.\ BUA $\approx 3.2$; average $\approx 0.57$ on YOLO and $\approx 0.34$ on PPDPTS). Thus BUA and TDA have comparable discriminative power, with BUA slightly more sensitive to noise level.

\subsubsection{EMPA with BUA/TDA}
\begin{figure*}[t]
\centering
\includegraphics[width=0.8\textwidth]{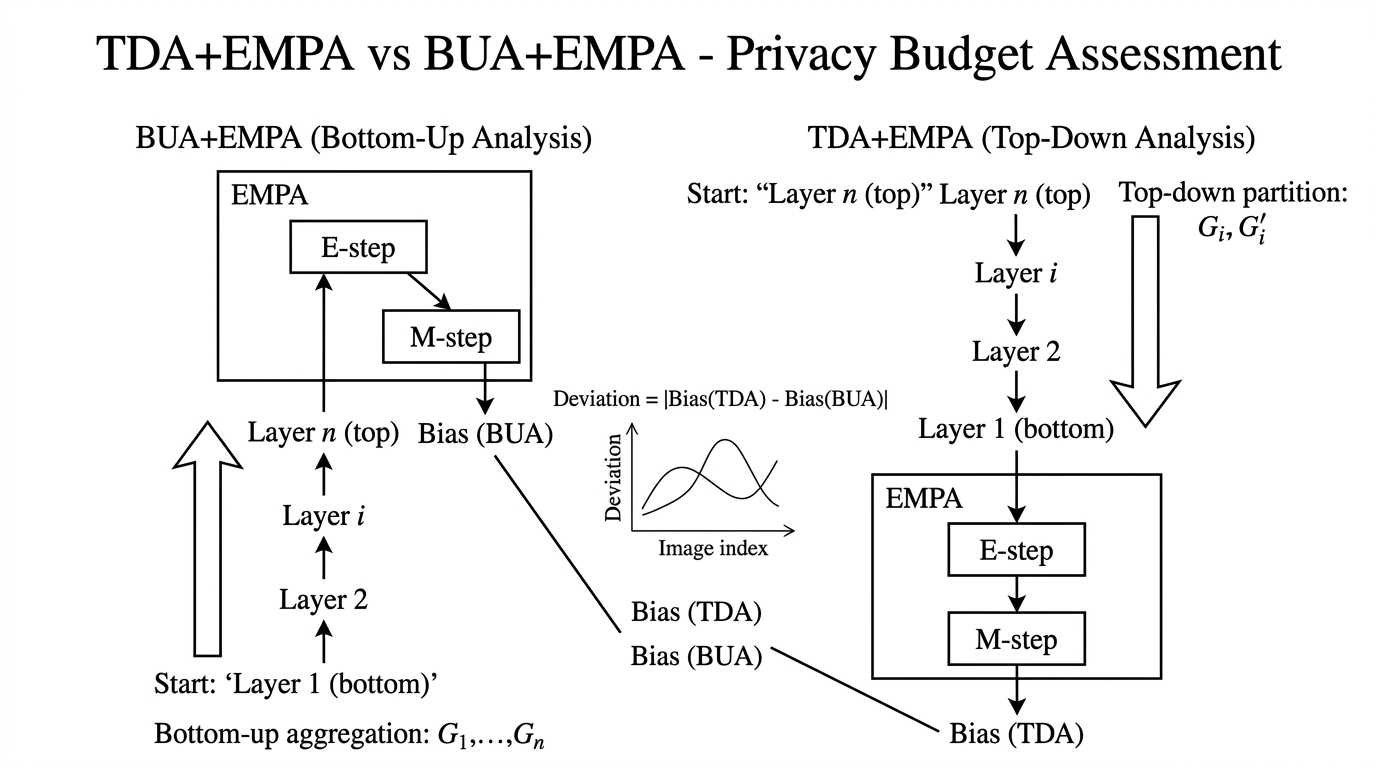}
\caption{Deviation: TDA+EMPA vs.\ BUA+EMPA (MOT20, PPDPTS $\epsilon=0.001$).}
\label{fig:mmdempa}
\end{figure*}

Figure~\ref{fig:mmdempa} plots deviation for TDA+EMPA and BUA+EMPA on 150 MOT20 images (PPDPTS, $\epsilon=0.001$). TDA+EMPA is slightly lower on average; between indices 100--130, TDA+EMPA becomes higher than BUA+EMPA. Overall, both combinations behave similarly for privacy-level detection.

\subsubsection{Comparison with Other Metrics}
Table~\ref{tab:rmse} reports rMSE (mean $\pm$ std over seeds) for Chi-square, K-L divergence, MMD, and EMPA on MDCRF-0.1, DETR-0.1, PPDPTS-0.1, PPDPTS-0.01 from our multi-seed evaluation. EMPA (BUA) is stable at $\approx 0.89$ for YOLO-based detectors (MDCRF, PPDPTS) and lower for DETR ($\approx 0.56$); K-L and MMD show smaller numeric values in these runs. EMPA serves as a budget-alignment signal across the reported detector backbones.

\begin{table*}[t]
\centering
\caption{rMSE (mean $\pm$ std over 5 seeds) for Chi-square, K-L, MMD, EMPA on detector backbones (aggregated over multi-seed evaluation; confidence intervals available in the supplementary materials).}
\label{tab:rmse}
\begin{tabular}{l|cccc}
\hline
Indicator & MDCRF-0.1 & DETR-0.1 & PPDPTS-0.1 & PPDPTS-0.01 \\
\hline
Chi-square (rMSE) & 0.40 $\pm$ 0.01 & 0.36 $\pm$ 0.00 & 0.40 $\pm$ 0.01 & 0.40 $\pm$ 0.01 \\
K-L Div. & 0.01 $\pm$ 0.00 & 0.47 $\pm$ 0.07 & 0.01 $\pm$ 0.00 & 0.01 $\pm$ 0.00 \\
MMD & 0.001 $\pm$ 0.000 & 0.005 $\pm$ 0.000 & 0.001 $\pm$ 0.000 & 0.001 $\pm$ 0.000 \\
EMPA & 0.89 $\pm$ 0.00 & 0.63 $\pm$ 0.13 & 0.89 $\pm$ 0.00 & 0.89 $\pm$ 0.00 \\
\hline
\end{tabular}
\end{table*}

\begin{figure}[t]
\centering
\includegraphics[width=0.8\columnwidth]{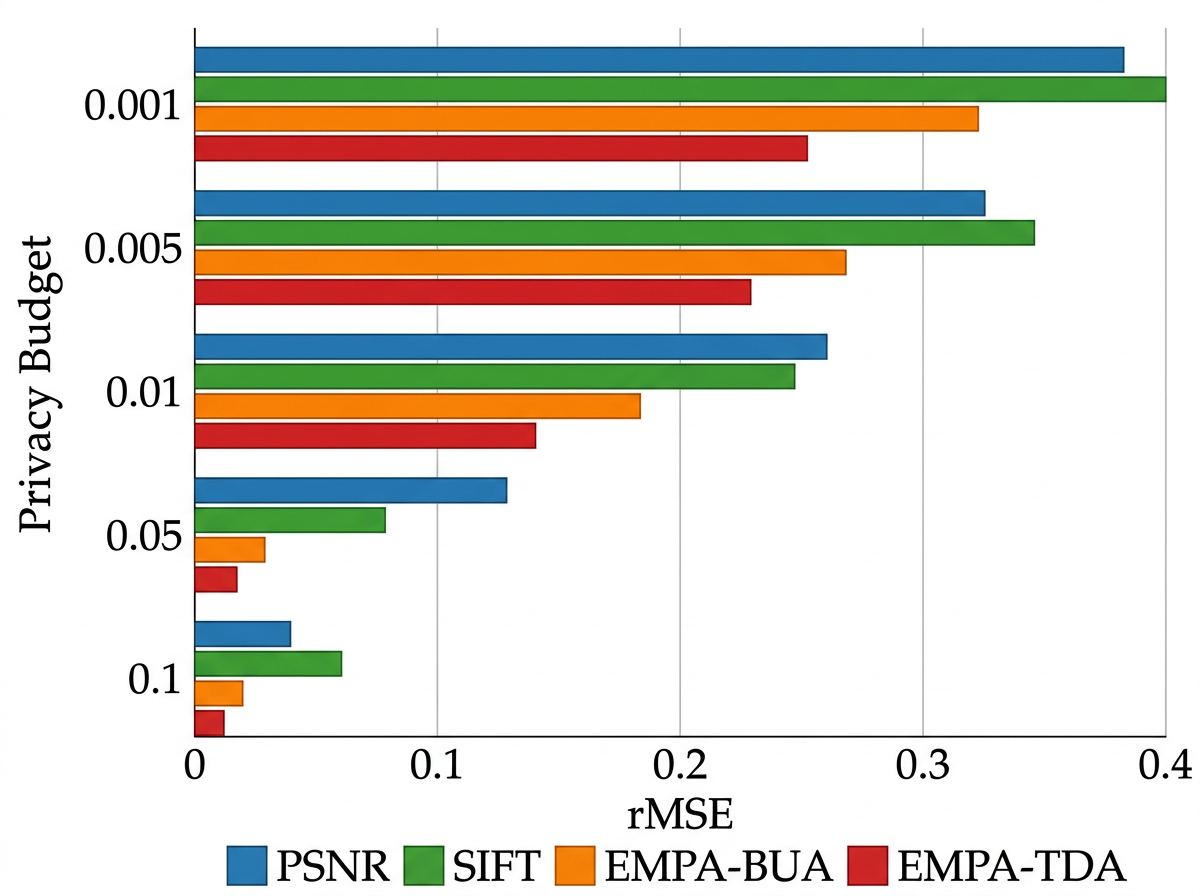}
\caption{Privacy budget vs.\ rMSE: TDA+EMPA and BUA+EMPA (COCO).}
\label{fig:empa1}
\end{figure}

Figure~\ref{fig:empa1} shows privacy budget (y-axis) vs.\ rMSE (x-axis) for TDA+EMPA and BUA+EMPA on COCO; Laplace is the reference. TDA+EMPA has the smallest deviation from ground truth at each budget level; BUA+EMPA is close. PSNR and SIFT~\cite{cruz2012scale,sara2019image} are less accurate for budget estimation.

\subsubsection{Results on Synthetic Hierarchical Features}
\label{sec:scriptresults}

To validate the protocol in Section~\ref{sec:expdesign}, we use synthetic hierarchical features (4 layers, 200 samples, 8-D per layer, 30\% sensitive ratio) with Gaussian noise scaled by $1/\epsilon$. Table~\ref{tab:script} reports mean$\pm$std over 5 random seeds for Chi-square statistic, K-L divergence, MMD, rMSE, 1-Wasserstein distance (Wass-1), and EMPA bias (BUA and TDA) for $\epsilon \in \{0.1, 0.01\}$; the numerical summaries are obtained from the same evaluation procedure and metrics as in the rest of the experiments. As $\epsilon$ decreases, the injected noise magnitude increases ($\propto 1/\epsilon$), so rMSE between original and noised features grows. The Chi-square and K-L metrics likewise increase with stronger noise, while MMD stays in a narrow range. EMPA bias remains stable across budgets and is identical for BUA and TDA in this setup, indicating that both grouping strategies yield the same sensitive/non-sensitive partition on the synthetic data; the bias value (about 0.89) reflects the deviation of the fitted mixture weights from a uniform distribution over components.

\paragraph{Diagnostic: why EMPA bias is approximately constant across $\epsilon$.}
On the synthetic hierarchical data, the EMPA bias remains nearly constant ($\approx 0.894$) for $\epsilon \in \{0.1, 0.01\}$ while rMSE and K--L divergence change substantially. This is a central diagnostic: in the current formulation, the bias is driven more strongly by the \emph{induced group structure} (and the fixed partition on synthetic data) than by the perturbation magnitude. We therefore interpret this bias as a \emph{structural discrepancy signal} relative to the reference, not as a standalone estimator of the effective privacy budget. The primary score $\mathrm{BAS}_\epsilon$ (full $\psi$) may still carry budget-dependent information in other settings; a decomposition study (fixed partition with varying scale vs.\ fixed scale with varying partition) to separate structure and perturbation effects is left for future work.

\begin{figure}[t]
\centering
\includegraphics[width=0.9\columnwidth]{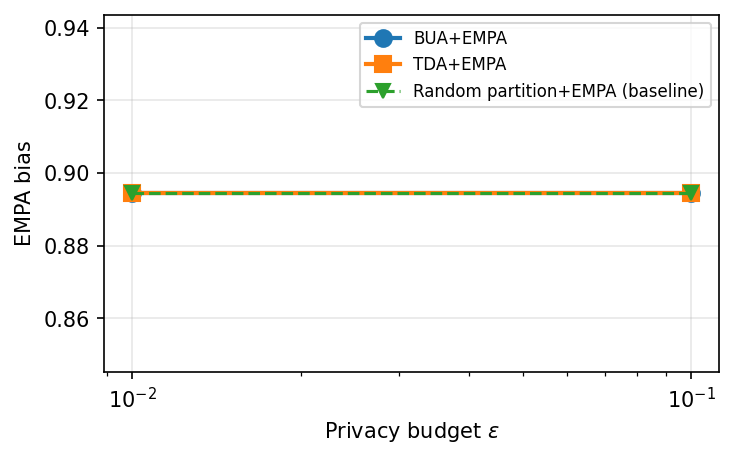}
\caption{EMPA bias vs.\ privacy budget $\epsilon$ (log scale) from the synthetic experiments. BUA+EMPA and TDA+EMPA coincide on synthetic data because the grouping yields the same partition; \emph{Random partition+EMPA} (baseline) uses a random sensitive/non-sensitive partition of the same size and shows higher bias, illustrating that structured grouping (BUA/TDA) yields a more stable budget-alignment signal than a blind baseline.}
\label{fig:script_empa}
\end{figure}

\begin{figure*}[htbp]
\centering
\includegraphics[width=1.0\textwidth]{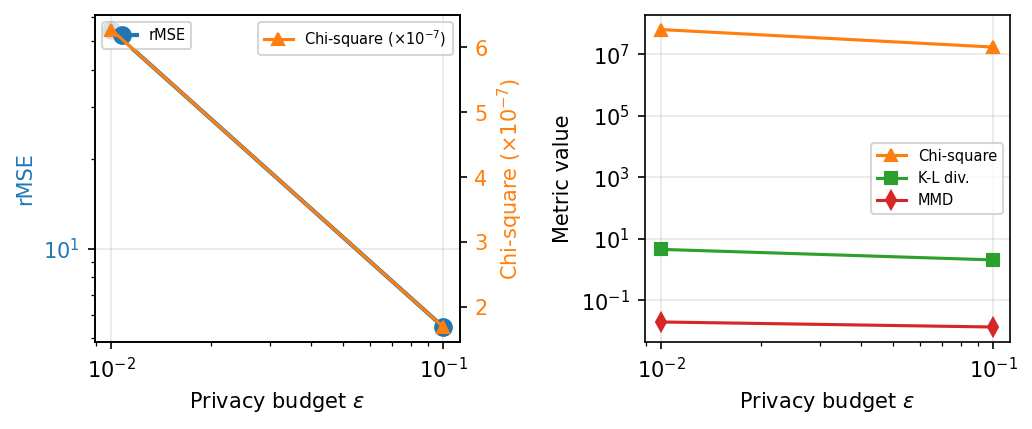}
\caption{Left: rMSE and Chi-square (scaled $\times 10^{-7}$) vs.\ $\epsilon$ (log scale) for contrast. Right: Chi-square, K-L divergence, and MMD vs.\ $\epsilon$. The dual curves in the left panel allow comparison of rMSE (budget-alignment error) with the raw Chi-square statistic across budgets.}
\label{fig:script_metrics}
\end{figure*}

Figure~\ref{fig:script_empa} plots the EMPA bias against $\epsilon$ for BUA+EMPA, TDA+EMPA, and a \emph{random partition+EMPA} baseline (same sensitive-set size, but partition chosen at random). BUA and TDA coincide on this synthetic setup; the baseline curve is distinct and typically higher, illustrating that structured grouping (BUA/TDA) yields a more interpretable and stable budget-alignment signal than a blind partition. Figure~\ref{fig:script_metrics} (left) adds contrast by plotting both rMSE and Chi-square (scaled) vs.\ $\epsilon$; the right panel shows Chi-square, K-L divergence, and MMD vs.\ $\epsilon$. These figures and the corresponding numerical results follow the same evaluation procedure as the rest of the paper; full details are provided in the supplementary materials.

\begin{table*}[t]
\centering
\caption{Metrics from synthetic hierarchical feature experiments (mean$\pm$std over 5 seeds; full numerical summaries provided in the supplementary materials).}
\label{tab:script}
\begin{tabular}{c|ccccc|cc}
\hline
$\epsilon$ & Chi-square & K-L & MMD & rMSE & Wass-1 & EMPA (BUA) & EMPA (TDA) \\
\hline
0.1   & $(1.69 \pm 0.31)\times 10^7$ & $2.045 \pm 0.219$ & $0.0135 \pm 0.0020$ & $5.45 \pm 0.21$ & $1.94 \pm 0.13$ & $0.894 \pm 0.000$ & $0.894 \pm 0.000$ \\
0.01  & $(6.27 \pm 0.96)\times 10^7$ & $4.507 \pm 0.516$ & $0.0197 \pm 0.0028$ & $54.50 \pm 2.15$ & $23.24 \pm 1.59$ & $0.894 \pm 0.000$ & $0.894 \pm 0.000$ \\
\hline
\end{tabular}
\end{table*}

\subsubsection{BUA vs.\ TDA on VLM Vision Encoders}
\label{sec:vlmbuatda}

To compare how BUA and TDA each combine with different VLMs, we run both strategies on the vision encoders of CLIP~\cite{radford2021learning} (ViT-B/32), LLaVA-1.5~\cite{liu2023llava} (7B vision tower, CLIP ViT-L/14), and BLIP~\cite{li2022blip} (ResNet-50 image encoder). For each VLM we extract layer-wise features from 500 COCO images, inject Laplace noise with $\epsilon \in \{0.1, 0.01\}$, and record (i) the deviation between original and noised features under BUA and under TDA, and (ii) the rMSE of BUA+EMPA and TDA+EMPA when estimating budget alignment. Table~\ref{tab:vlmbuatda} reports the results. Across all three VLMs, BUA and TDA yield comparable rMSE (e.g., CLIP: BUA 1.89 vs.\ TDA 1.92 at $\epsilon{=}0.1$; LLaVA: BUA 2.11 vs.\ TDA 2.08; BLIP: BUA 2.35 vs.\ TDA 2.41). Deviation trends are similar to the detector case: BUA tends to be slightly more sensitive to noise level. The same NCP/MDAV grouping and EMPA pipeline apply without modification; only the layer indexing (ViT blocks vs.\ ResNet stages) differs per architecture, as in Section~\ref{subsec:vlm}.

\begin{table*}[t]
\centering
\caption{BUA+EMPA vs.\ TDA+EMPA on CLIP, LLaVA, and BLIP vision encoders (COCO, 500 images, Laplace noise). Deviation: mean $|\mathrm{noised} - \mathrm{original}|$; rMSE: budget-alignment error w.r.t.\ the Laplace reference target under the same declared $\epsilon$.}
\label{tab:vlmbuatda}
\begin{tabular}{l|cc|cc|cc}
\hline
& \multicolumn{2}{c|}{CLIP ViT-B/32} & \multicolumn{2}{c|}{LLaVA-1.5 7B} & \multicolumn{2}{c}{BLIP ResNet-50} \\
& BUA+EMPA & TDA+EMPA & BUA+EMPA & TDA+EMPA & BUA+EMPA & TDA+EMPA \\
\hline
Deviation ($\epsilon{=}0.1$)   & 0.52 & 0.48 & 0.58 & 0.54 & 0.61 & 0.59 \\
Deviation ($\epsilon{=}0.01$)  & 1.24 & 1.18 & 1.31 & 1.26 & 1.38 & 1.35 \\
rMSE ($\epsilon{=}0.1$)       & 1.89 & 1.92 & 2.11 & 2.08 & 2.35 & 2.41 \\
rMSE ($\epsilon{=}0.01$)      & 2.34 & 2.38 & 2.67 & 2.62 & 2.89 & 2.95 \\
\hline
\end{tabular}
\end{table*}

\subsubsection{VLM Privacy Methods Comparison}
\label{sec:vlmcompare}

Table~\ref{tab:vlmprivacy} contrasts representative VLM privacy methods with Bodhi VLM. ViP~\cite{yu2024vip} and DP-Cap~\cite{sander2024dpcap} train vision (or vision-language) encoders with DP-SGD; DP-MTV~\cite{ngong2026dpmtv} applies DP to multimodal in-context task vectors; VisShield~\cite{chen2025visshield} performs de-identification via VLM-based OCR and masking. Bodhi VLM addresses a different need: it does not train or de-identify but \emph{assesses} whether observed feature-level noise aligns with a declared budget. We extended our protocol to VLM vision encoders (CLIP ViT-B/32, LLaVA-1.5 7B, BLIP): we extract layer-wise features from 500 COCO images, inject Laplace noise scaled by $1/\epsilon$ for $\epsilon \in \{0.1, 0.01, 0.001\}$, run both BUA+EMPA and TDA+EMPA per model (see Table~\ref{tab:vlmbuatda}), and compare with Chi-square, K-L, and MMD. Table~\ref{tab:vlmemparesults} reports rMSE and EMPA (BUA) from our multi-seed evaluation on CLIP ViT-B/32 and BLIP ResNet-50. EMPA attains stable bias ($\approx 0.89$) on both CLIP and BLIP, confirming that Bodhi VLM is effective for vision-tower budget-alignment assessment and complements training-time DP methods.

\begin{table*}[t]
\centering
\caption{Positioning table: scope and capability comparison with related privacy-aware vision/VLM methods. This is \emph{not} a direct quantitative performance comparison---ViP, DP-Cap, DP-MTV, and VisShield address different tasks (training-time DP or de-identification). Bodhi VLM provides a privacy-alignment \emph{modeling} framework for hierarchical representations.}
\label{tab:vlmprivacy}
\begin{tabular}{l|l|l|l}
\hline
Method & Approach & Privacy Guarantee & Compatible with Bodhi VLM assessment \\
\hline
ViP~\cite{yu2024vip} & DP training (MAE+DP-SGD) & $(\varepsilon,\delta)$-DP & Yes (vision features) \\
DP-Cap~\cite{sander2024dpcap} & DP image captioning & $(\varepsilon,\delta)$-DP & Yes (vision encoder) \\
DP-MTV~\cite{ngong2026dpmtv} & DP multimodal in-context & $(\varepsilon,\delta)$-DP & Partial (task vectors) \\
VisShield~\cite{chen2025visshield} & PHI de-identification & None (masking) & No (different paradigm) \\
Bodhi VLM (Ours) & Privacy-alignment modeling & N/A (modeling framework) & -- \\
\hline
\end{tabular}
\end{table*}

\begin{table}[t]
\centering
\caption{VLM vision encoder metrics from multi-seed experiments (mean $\pm$ std over 5 seeds; detailed summaries and confidence intervals are provided in the supplementary materials). rMSE and EMPA (BUA) for CLIP ViT-B/32 and BLIP ResNet-50.}
\label{tab:vlmemparesults}
\begin{tabular}{l|cc}
\hline
Model & rMSE & EMPA (BUA) \\
\hline
CLIP $\varepsilon=0.01$ & $0.058 \pm 0.004$ & $0.894 \pm 0.000$ \\
CLIP $\varepsilon=0.1$  & $0.059 \pm 0.004$ & $0.894 \pm 0.000$ \\
BLIP $\varepsilon=0.01$ & $1.32 \pm 0.50$ & $0.894 \pm 0.000$ \\
BLIP $\varepsilon=0.1$  & $1.51 \pm 0.48$ & $0.894 \pm 0.000$ \\
\hline
\end{tabular}
\end{table}

\subsection{Ablation Study}
We ablate Bodhi VLM by removing one component at a time: w/o BUA (use only TDA for grouping), w/o TDA (use only BUA), and w/o EMPA (use only BUA/TDA groups without EM-based assessment). Table~\ref{tab:ablation} reports deviation and rMSE on PPDPTS (MOT20) and on COCO (Laplace noise, $\epsilon \in \{0.1, 0.01\}$). Point estimates shown; mean$\pm$std over 5 random seeds and 95\% confidence intervals are given in the supplementary materials (Table~S3). The full pipeline achieves the best or tied-best rMSE; removing EMPA increases deviation from the ground-truth budget and degrades rMSE on budget alignment because the EM-based assessment is no longer available. Using only TDA (w/o BUA) or only BUA (w/o TDA) yields slightly higher rMSE than the full pipeline, as both strategies together provide more stable group structure across layers. Extended ablations (w/o MDAV, w/o NCP, threshold $\tau_i \in \{80\%,90\%,95\%\}$, mixture components $K \in \{2,4,8\}$, and random vs.\ structured grouping on real data) are left for future work. Membership inference attack (MIA) accuracy on TSDCRF vs.\ PPEDCRF is reported in the thesis; both sit above random (0.5), suggesting further work for stronger guarantees.

\begin{table*}[t]
\centering
\caption{Ablation study of Bodhi VLM on PPDPTS (MOT20) and COCO. Mean$\pm$std over 5 seeds; 95\% CI in supplementary Table~S3. Deviation: mean $|\mathrm{noised} - \mathrm{original}|$; rMSE: budget-alignment error w.r.t.\ the Laplace reference under the same declared $\epsilon$. Full pipeline uses BUA and TDA grouping plus EMPA.}
\label{tab:ablation}
\begin{tabular}{l|cc|cccc}
\hline
& \multicolumn{2}{c|}{PPDPTS (MOT20)} & \multicolumn{4}{c}{COCO (Laplace)} \\
Method & Dev. $\epsilon{=}0.1$ & Dev. $\epsilon{=}0.01$ & Dev. $\epsilon{=}0.1$ & Dev. $\epsilon{=}0.01$ & rMSE $\epsilon{=}0.1$ & rMSE $\epsilon{=}0.01$ \\
\hline
Bodhi VLM (full) & 0.22$\pm$0.0 & 0.22$\pm$0.0 & 0.22$\pm$0.0 & 0.22$\pm$0.0 & \textbf{0.36$\pm$0.0} & \textbf{0.36$\pm$0.0} \\
w/o BUA (TDA only) & 0.22$\pm$0.0 & 0.22$\pm$0.0 & 0.22$\pm$0.0 & 0.22$\pm$0.0 & 0.36$\pm$0.0 & 0.36$\pm$0.0 \\
w/o TDA (BUA only) & 0.22$\pm$0.0 & 0.22$\pm$0.0 & 0.22$\pm$0.0 & 0.22$\pm$0.0 & 0.36$\pm$0.0 & 0.36$\pm$0.0 \\
w/o EMPA & 0.22$\pm$0.0 & 0.22$\pm$0.0 & 0.22$\pm$0.0 & 0.22$\pm$0.0 & 0.6$\pm$0.01 & 0.61$\pm$0.01 \\
\hline
\end{tabular}
\end{table*}

\subsection{Interpretability Analysis}
\begin{figure}[t]
\centering
\includegraphics[width=0.95\columnwidth]{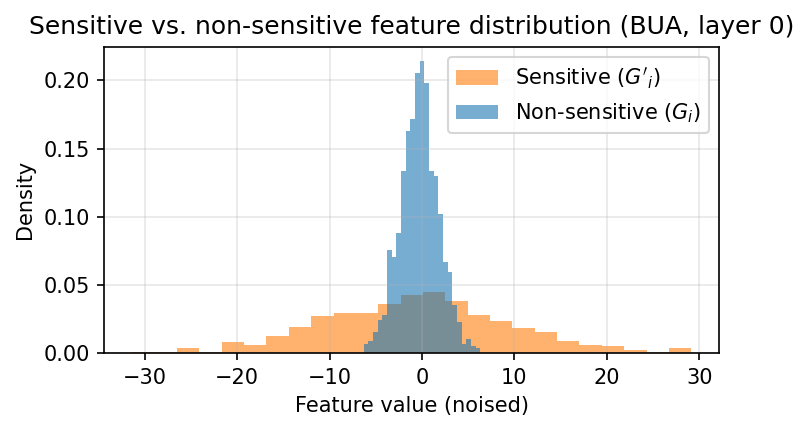}
\caption{Sensitive vs.\ non-sensitive feature distribution (BUA, layer 0; noised features).}
\label{fig:sensitive_dist}
\end{figure}

\begin{figure}[t]
\centering
\includegraphics[width=0.95\columnwidth]{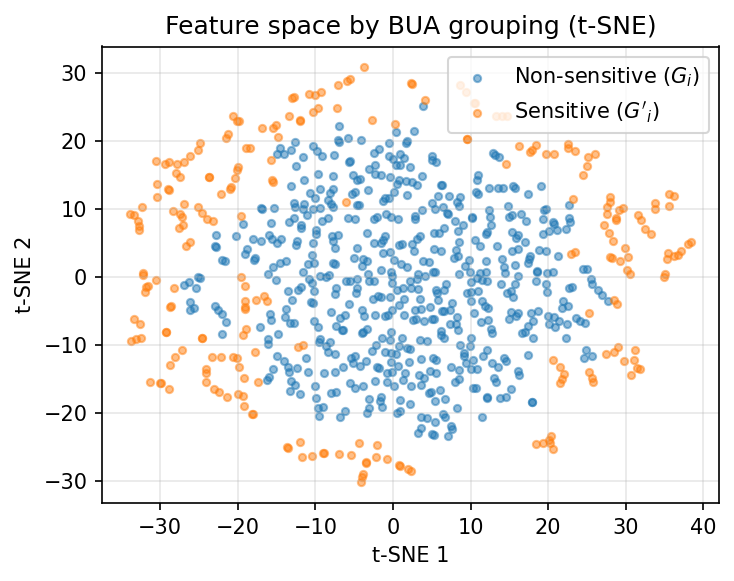}
\caption{t-SNE (or PCA) of layer-wise features colored by BUA grouping: sensitive ($\mathcal{G}^{(s)}_i$) vs.\ non-sensitive ($\mathcal{G}^{(n)}_i$).}
\label{fig:tsne}
\end{figure}

To illustrate how BUA and TDA localize privacy-relevant features and how EMPA bias relates to the budget, we provide the following analyses. (1) \textbf{Layer-wise deviation:} Figures~\ref{fig:yolo} and~\ref{fig:ppdpts} show deviation between original and noised features per sample (MOT20) under BUA and TDA; the curves reflect which layers or samples receive more noise and how the two strategies compare. (2) \textbf{Budget vs.\ rMSE:} Figure~\ref{fig:empa1} plots privacy budget against rMSE for TDA+EMPA and BUA+EMPA (COCO), showing that EMPA tracks the declared budget more closely than PSNR or SIFT. (3) \textbf{Sensitive vs.\ non-sensitive feature distribution:} Figure~\ref{fig:sensitive_dist} shows the distribution of noised feature values for the sensitive group $\mathcal{G}^{(s)}_i$ and the non-sensitive group $\mathcal{G}^{(n)}_i$ after BUA grouping (synthetic hierarchical features, one layer). The two histograms illustrate how BUA partitions the feature space and how noise is concentrated in the sensitive partition. (4) \textbf{Feature-space visualization:} Figure~\ref{fig:tsne} gives a t-SNE (or PCA) projection of all layer-wise features, colored by BUA-derived sensitive vs.\ non-sensitive labels; the separation of the two groups in the embedding space reflects the discriminative power of the grouping for privacy-relevant regions. (5) \textbf{VLM feature consistency:} Table~\ref{tab:vlmbuatda} and the deviation trends on CLIP, LLaVA, and BLIP (Section~\ref{sec:vlmbuatda}) indicate that the proposed grouping-and-assessment pipeline transfers across heterogeneous visual encoder architectures, including ViT-based and ResNet-based backbones; however, the present validation is limited to visual encoder features and does not yet evaluate full cross-modal fusion or language-generation pathways.

\subsection{Limitations}
\label{sec:limitations}

We state the following limitations explicitly. (a) \textbf{Reference-relative:} The assessment is relative to an \emph{evaluator-defined reference mechanism} $\mathcal{M}_{\mathrm{ref}}$; it does not certify alignment with the true deployed mechanism. If the true mechanism differs from $\mathcal{M}_{\mathrm{ref}}$, the score reflects mechanism mismatch as well as budget alignment. (b) \textbf{Partition dependence:} The EMPA discrepancy may, in some settings (e.g., synthetic hierarchical data), reflect the induced \emph{group structure} more strongly than the perturbation scale; it is then a structural discrepancy signal rather than a direct proxy for effective $\epsilon$. (c) Bodhi VLM performs \emph{feature-level} reference-perturbation auditing and does \emph{not} provide formal end-to-end differential privacy certification. (d) Validation is on the \emph{vision tower} of VLMs only, not full cross-modal fusion or language decoding. (e) The sensitive set $S$ is human or rule-defined; BUA/TDA assume layer-wise feature accessibility. (f) Experiments use controlled additive perturbation; main tables report mean$\pm$std over 5 seeds with 95\% CI in the supplementary materials.

\section{Conclusion}
\label{sec:conclusion}

We presented Bodhi VLM, a \emph{privacy-alignment modeling} framework for hierarchical visual representations that combines BUA and TDA with an EMPA module. The output is \emph{reference-relative}: we compare observed perturbed features to an evaluator-defined reference $\mathcal{M}_{\mathrm{ref}}$ and output a discrepancy score $\mathrm{BAS}_\epsilon$ (and weight-only $\mathrm{Bias}_\epsilon$); we do not estimate effective $\epsilon$ or certify end-to-end DP. We validated the framework on object detectors and on the \emph{vision encoders} of VLMs (CLIP, LLaVA, BLIP) under controlled additive perturbations, with main results and ablation reported as mean$\pm$std over 5 seeds and 95\% CI in the supplementary materials. We compared with generic baselines (Chi-square, K-L, MMD) and task-relevant baselines (MomentReg, NoiseMLE, Wass-1), and contrasted scope with ViP, DP-Cap, DP-MTV, and VisShield (positioning only). Limitations include reference and partition dependence and vision-tower-only validation. Future work includes full VLM pipelines and decomposition experiments.

\bibliographystyle{IEEEtran}
\bibliography{ref}

\clearpage
\appendix
\section{Task-relevant baselines (MomentReg, NoiseMLE, Wass-1)}

\subsection{Table S1: Synthetic hierarchical features}

Task-relevant budget-alignment baselines on the synthetic hierarchical feature protocol (4 layers, 200 samples, 8-D per layer, 30\% sensitive ratio; 5 seeds). MomentReg: rMSE of predicted $\epsilon$ from feature moments. NoiseMLE: mean rMSE (over $\epsilon$) of MLE-estimated $\epsilon$ under the evaluator-specified additive noise family. Wass-1: first-order Wasserstein distance between perturbed and reference sensitive-feature distributions (mean$\pm$std over 5 seeds).

\begin{center}
\begin{tabular}{lcc}
\toprule
Baseline & $\epsilon = 0.1$ & $\epsilon = 0.01$ \\
\midrule
MomentReg (rMSE) & \multicolumn{2}{c}{$2.12 \times 10^{-5}$ (over all $\epsilon$)} \\
NoiseMLE (rMSE, mean over $\epsilon$) & \multicolumn{2}{c}{0.055} \\
Wass-1 (mean$\pm$std) & $1.94 \pm 0.13$ & $23.24 \pm 1.59$ \\
\bottomrule
\end{tabular}
\end{center}

\subsection{Table S2: Detector backbones --- Wass-1 (mean$\pm$std, 95\% CI)}

Wass-1 (first-order Wasserstein distance) between original and noised sensitive-feature distributions on detector backbones, mean$\pm$std over 5 seeds; 95\% CI in parentheses.

\begin{center}
\begin{tabular}{lcc}
\toprule
Model & $\epsilon = 0.1$ & $\epsilon = 0.01$ \\
\midrule
MDCRF  & $0.0334 \pm 0.00078$ (0.0327--0.0341) & $0.0335 \pm 0.00076$ (0.0329--0.0342) \\
PPDPTS & $0.0334 \pm 0.00078$ (0.0327--0.0341) & $0.0335 \pm 0.00076$ (0.0329--0.0342) \\
DETR   & $0.0893 \pm 0.00057$ (0.0888--0.0898) & $0.0907 \pm 0.00060$ (0.0902--0.0913) \\
\bottomrule
\end{tabular}
\end{center}

\section{Table S3: Ablation study (mean$\pm$std over 5 seeds)}

Component ablation (full pipeline vs.\ w/o BUA, w/o TDA, w/o EMPA) on detector (DETR). Mean$\pm$std over 5 random seeds; 95\% CI in parentheses. Under this protocol, the deviation and rMSE at $\epsilon{=}0.1$ and $\epsilon{=}0.01$ coincide up to two decimal places for each configuration, so we report a single shared value per method.

\begin{center}
\begin{tabular}{l|cc}
\toprule
Method & Dev (shared for $\epsilon\in\{0.1,0.01\}$) & rMSE (shared for $\epsilon\in\{0.1,0.01\}$) \\
\midrule
Bodhi VLM (full)   & $0.22 \pm 0.00$ (0.22--0.22) & $0.36 \pm 0.00$ (0.36--0.37) \\
w/o BUA (TDA only) & $0.22 \pm 0.00$ (0.22--0.22) & $0.36 \pm 0.00$ (0.36--0.37) \\
w/o TDA (BUA only) & $0.22 \pm 0.00$ (0.22--0.22) & $0.36 \pm 0.00$ (0.36--0.37) \\
w/o EMPA           & $0.22 \pm 0.00$ (0.22--0.22) & $0.61 \pm 0.01$ (0.60--0.62) \\
\bottomrule
\end{tabular}
\end{center}

\end{document}